\documentclass[aps,pre,twocolumn,amsmath,amssymb,showpacs]{revtex4}
\usepackage{epsfig}
\begin{document}
\title{Emotional Analysis of Blogs and Forums Data}
\author{Pawe{\l} Wero\'{n}ski$^a$, Julian Sienkiewicz$^a$, Georgios Paltoglou$^b$, Kevan Buckley$^b$, Mike Thelwall$^b$ and Janusz A. Ho{\l}yst$^{a,}$ \footnote{corresponding author, email: jholyst@if.pw.edu.pl}}
\affiliation{$^a$Faculty of Physics, Center of Excellence for Complex Systems Research, Warsaw University of Technology, Koszykowa 75, PL-00-662 Warsaw, Poland.\\
$^b$Statistical Cybermetrics Research Group, School of Technology, University of Wolverhampton, Wulfruna Street, WV1 1LY Wolverhampton, United Kingdom}
\begin{abstract}
We perform a statistical analysis of emotionally annotated comments in two large online datasets, examining chains of consecutive posts in the discussions. Using comparisons with randomised data we show that there is a high level of correlation for the emotional content of messages.  
\end{abstract}
\maketitle

\section{Introduction}	
Recent years have resulted in several well motivated and carefully described studies coping with the problem of opinion formation and its spreading \cite{castellano}. This kind of research usually aimed at qualitative descriptions of some specific phenomena using both numerical and analytical methods and touched problems like culture dissemination \cite{axelrod}, decision making \cite{sznajd-weron}, majority rule voting \cite{galam}, social impact \cite{kacperski} or community isolation \cite{sienkiewicz}. The bottleneck of such studies is always the lack of real-world data that could sustain the presented theories. On the other hand, the rapid and overwhelming development of the Internet enables gathering information on its users and their habits, spotting characteristic structures \cite{bernard} and users' behaviour \cite{anka}. However, all these works have not delved into a crucial aspect of any analysis of new-born media like Internet blogs or forums: their emotional content. It is only lately that such analyses have started to emerge \cite{hate,bosa1,bosa2,frank}. In this paper we focus on the properties of emotionally annotated chains of posts from two large online datasets. We give strong evidence that the discussions cannot be treated as random insertions of comments showing various measures of correlation with the emotional content of posts. The paper is complementary to recent analyses of cluster formation and the influence of negative emotions on the properties of online discussions \cite{ania_plos,ania_physa}.

\section{Data description}	
The aim of this study was to find common properties of comment chains in Internet blogs. The analysis was performed on two datasets: Blogs and BBC Forums. The BBC web site had a number of publicly-open moderated Message Boards covering a wide variety of topics that allow registered users to start their own discussions and post comments on existing discussions. Our data included discussions posted on the Religion and Ethics and World/UK News message boards starting from the launch of the website (July 2005 and June 2005 respectively) until June 2009. The Blogs dataset is a subset of the Blogs06 \cite{blogs06} collection of blog posts from 06/12/2005 to 21/02/2006. Only posts attracting more than 100 comments were extracted, as these apparently initialised non-trivial discussions. Both datasets have similar structures. They consist of blog posts and corresponding indexed comments possessing two values: positive probability $P_{pos}$ and subjective probability $P_{pos}$ (both are real numbers between 0 and 1) forming a chain of comments $x_1,x_2,...,x_{n-1},x_N$ ($N$ is the thread length). These values are the output of a sentiment analysis classifier that is informed by previous studies on the extraction of emotion from texts \cite{pang,mike}. Sentiment analysis algorithms often operate in stages as follows: (a) separating objective from subjective texts, (b) predicting the polarity of the subjective texts, and (c) detecting the sentiment target \cite{25}. Our algorithm used supervised, machine-learning principles \cite{28}. For this, we implemented an hierarchical extension of a standard Language Model (LM) classifier \cite{28}. LM classifiers estimate the probability that a given document belongs to each class and then select the class with the highest probability. In our hierarchical extension a document is first classified by the algorithm as objective or subjective and then, for subjective texts a second-stage classification determines the polarity as being either positive or negative. We used a manually annotated subset of about 34,000 documents from the Blogs06 data set as a training corpus. The processed datasets have the following key properties: Blogs consists of 1,232 discussions (threads) with 245,698 comments in total while the BBC Forums have 97,946 threads with 2,474,781 comments. 

\section{$P_{pos}$ and $P_{sub}$ distributions}	
\begin{figure}[!ht]
\centerline{\psfig{file=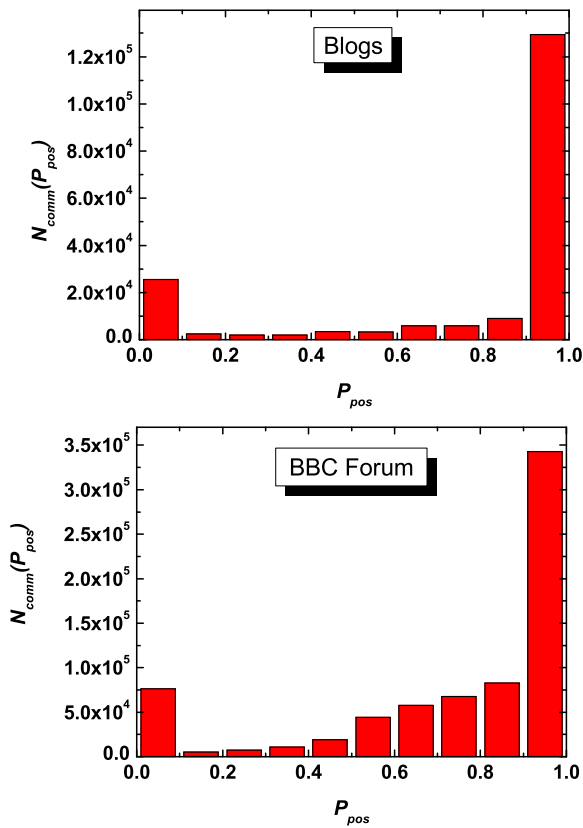,width=0.9\columnwidth}}
\caption{Histograms of positive probability values $P_{pos}$ for Blogs (upper plot) and BBC Forum (bottom plot).}
\label{fig:h1}
\end{figure}

\begin{figure}[!ht]
\centerline{\psfig{file=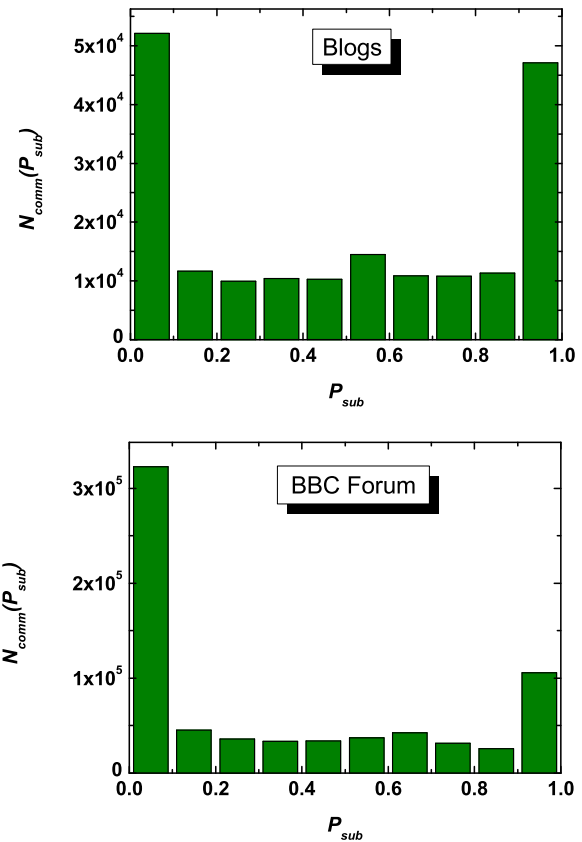,width=0.9\columnwidth}}
\caption{Histograms of subjective probability values $P_{sub}$ for Blogs (upper plot) and BBC Forum (bottom plot).}
\label{fig:h2}
\end{figure}

Histograms of $P_{pos}$ (Fig. \ref{fig:h1}) and $P_{sub}$ (Fig. \ref{fig:h2}) distributions were created. One can see that in both cases we could approximate the distributions by a bimodal distribution. In both cases there are two dominating histogram bars related to the extreme values 0 and 1. Therefore, by looking on Fig. 1 one could say that statistically very probably positive comments (later called positive) and very probably negative comments (later called negative) occur in threads in large quantities.
\section{Mean $\langle P_{pos} \rangle$ values in threads}	
\begin{figure}[!ht]
\centerline{\psfig{file=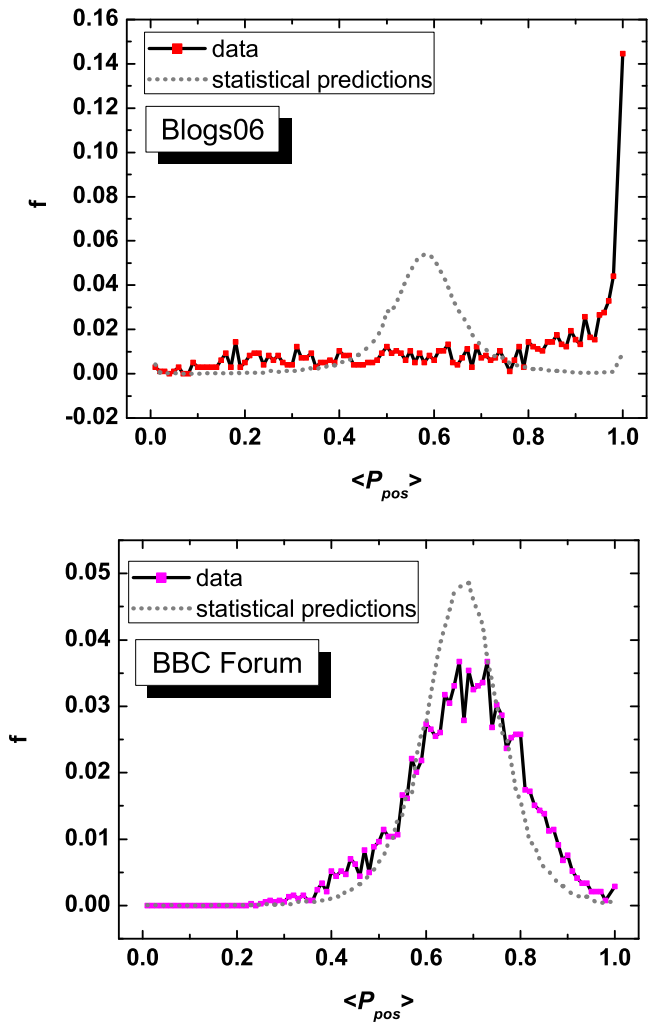,width=\columnwidth}}
\caption{Average positive probability frequency $f(P_{pos})$ for subjective comments in case of Blogs (upper plot) and BBC Forum (bottom plot). Solid lines with squares come from data and dotted lines are statistical predictions.}
\label{fig:posfreq}
\end{figure}

\begin{figure}[!ht]
\centerline{\psfig{file=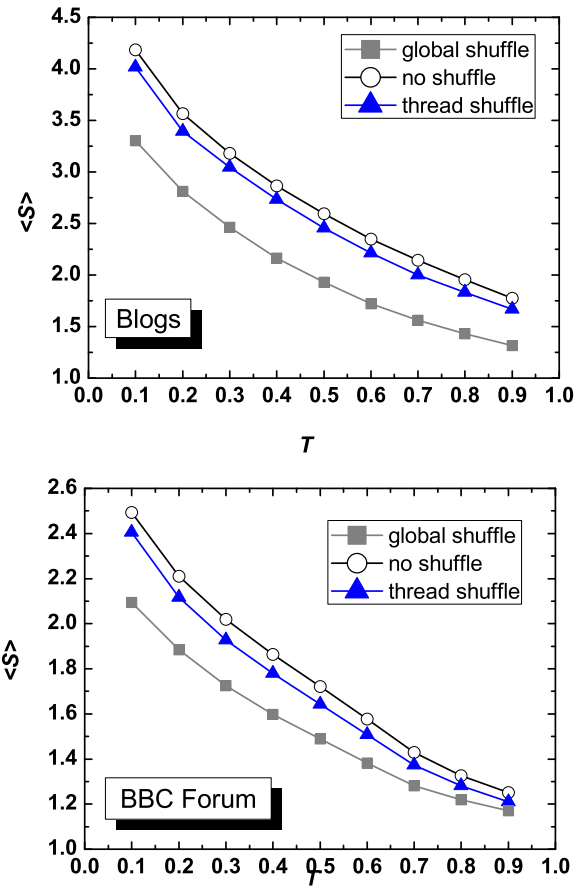,width=0.9\columnwidth}}
\caption{Average size of the subjective comments cluster $\langle S(T) \rangle$ for a given threshold $T$ for Blogs (upper plot) and BBC Forums (bottom plot). Circles represent data without shuffling, triangles come from shuffling at the thread level and squares represent data from global shuffling.}
\label{fig:clustsize}
\end{figure}

For each thread a mean value $\langle P_{pos} \rangle$ was calculated for all comments (Fig. \ref{fig:posfreq}). As a comparison, statistical predictions were used and every comment in each thread had its $P_{pos}$ randomised, using the $P_{pos}$ distribution (Fig. \ref{fig:h1}). One can see that in case of both datasets the plots of mean value distributions have a similar, Gaussian-like shape with a peak lower than the random predictions. For both datasets there is a shift toward positive values, which is much stronger in case of Blogs where the peak is centred in $P_{pos} = 1$. This difference between data statistics and statistical predictions of the shuffled data indicates the presence of strong correlations for positive comments in individual threads in the Blogs data. These correlations can be effects of mutual affective interactions between each thread's participants.
Any sequence of neighbouring comments within a thread which satsify the rule $P_{sub} \ge T$ would form a subjective cluster. For each threshold $T$ an average subjective cluster size $\langle S(T) \rangle$ was calculated (Fig. \ref{fig:clustsize}) and in order to make a reference {\it thread shuffling} and {\it global shuffling} were used. Here, shuffling is understood as a method of random reordering of the time series in order to destroy any existing correlations within the data. It can be performed at the thread level by randomising the $i$ index of the comment within a thread. Global shuffling is a process of randomising data within the whole dataset. In the case of the Blogs dataset the subjective attraction has a large impact on the structure of every comment chain (Fig. \ref{fig:clustsize}, upper plot). In comparison to global shuffling, the structure of the original data had clusters 26-36 \% greater in size. Another observation is be that in comparison to thread shuffling this increase was about 7 \%. It seems that some comment chains have a structure that can be only destroyed by global shuffling - for example, almost all posts being strongly positive or almost all being strongly negative. The main difference between the two datasets is that the BBC Forums  are less clustered (Fig. \ref{fig:clustsize}, bottom plot). This conclusion matches the much weaker subjectivity attraction behaviour of this dataset. The subjective probability in the $n$th comment is less likely to induce a similar value in the next comment, so it is common that more drastic shifts occur and therefore the clusters break more often. As a consequence of this, the mean cluster size is much smaller.

\section{Correlations for subjective probabilities}	
\begin{figure}[!h]
\centerline{\psfig{file=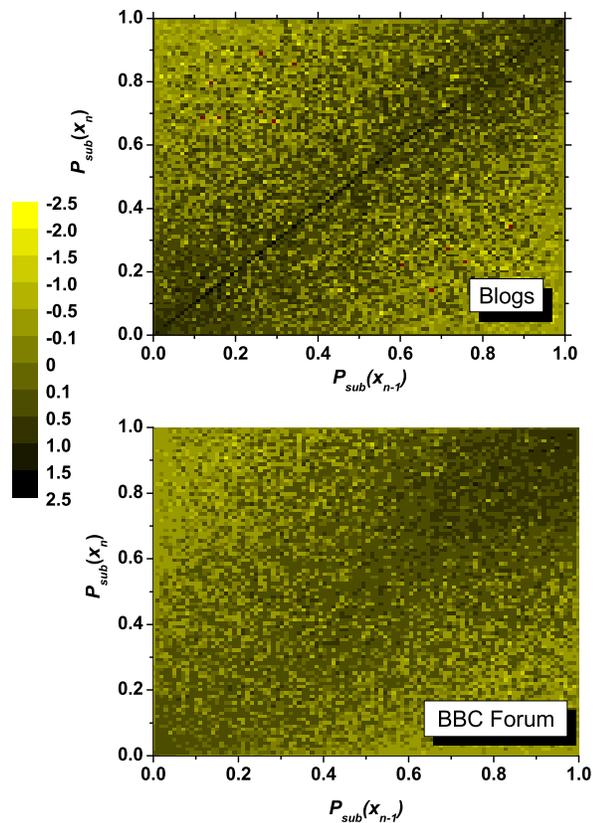,width=0.9\columnwidth}}
\caption{PMI for all pairs in all threads in case of $P_{sub}$ for Blogs (upper plot) and BBC Forum (bottom plot).}
\label{fig:PMI}
\end{figure}

In order to analyze the structure of comment grouping a probability correlation ratio was defined
\begin{equation}
C(x_n,x_{n-1})=\frac{p(x_n|x_{n-1})}{p(x_n)},
\label{eq:c}
\end{equation}
where $p(x_n|x_{n-1})$ is a conditional probability (here $x_n$ stands for $P_{sub}(n)$). The coefficient measures how the $(n-1)$-th state affects the $n$-th state in comparison to simply picking the $n$-th state at random. For instance $C=2$ would mean that in the analysed data subset the probability of getting $x_n$ if the previous comment was $x_{n-1}$ is two time greater than picking the $x_n$ value at random.  If the dataset is purely random in nature all the correlation ratio values would be equal to $C = 1$. The correlation ratio was calculated in the form of PMI (Pointwise Mutual Information) as $PMI = \log C$ in case of subjective probability for each pair in all threads (Fig. \ref{fig:PMI}). One can see that in case of both datasets the distribution values increase while closing to the diagonal, the $x_{n-1} = x_n$ line. This trend is very strong in case of Blogs (Fig. \ref{fig:PMI}, upper plot), the diagonal line is very distinct. For the BBC Forums there is also a more correlated area of the diagonal which lies in range $x_n \ge 0.7$ and $x_{n-1} \ge 0.7$.
\subsection{Mutual information}	
To quantify the amount of mutual dependence between two consecutive comments in the thread one can also use the concept of mutual information $I(X,Y)$ \cite{mutinfo}. It is formally defined for two discrete random variables $X$ and $Y$ as
\begin{equation}
I(X,Y) = \sum_{y \in Y}\sum_{x \in X}p(x,y)\log \frac{p(x,y)}{p(x)p(y)}
\end{equation}
where $p(x,y)$ is the joint probability function of $X$ and $Y$ while $p(x)$ and $p(y)$ are the marginal probability distribution functions of $X$ and $Y$. In our case random variable $X$ is equivalent to $P_{pos}$ (or $P_{sub}$) value of $n$-th comment while variable $Y$ is $P_{pos}$ (or $P_{sub}$) value of $(n-1)$-th comment. The results of the calculations are shown in Tables \ref{tab:pos} and \ref{tab:sub}. As one can see that the values obtained for Blogs are significantly different than those for globally reshuffled data. This suggests that $n$-th and $(n-1)$-th comments are not independent of each other. Similar results for BBC Forums are less pronounced, which is probably related to tree-like structure of those forums.

\begin{table}[ht]
\centerline{
\begin{tabular}{|c|c|c|c|c|c|}
  \hline
\multicolumn{2}{|c|}{No shuffle} & \multicolumn{2}{|c|}{Thread Shuffle} & \multicolumn{2}{|c|}{Global Shuffle}\\
  \hline
  \hline
  Blogs & BBC & Blogs & BBC & Blogs & BBC\\
  \hline
   4.53 & 0.41 & 3.57 & 0.26 & 0.05 & 0\\
  \hline
\end{tabular}}
\caption{Mutual information for positive probability value of subsequent comments. The value of $I(X,Y)$ was calculated with about 0.05 error due to calculation method simplifications}\label{tab:pos}
\end{table}

\begin{table}[ht]
\centerline{
\begin{tabular}{|c|c|c|c|c|c|}
  \hline
\multicolumn{2}{|c|}{No shuffle} & \multicolumn{2}{|c|}{Thread Shuffle} & \multicolumn{2}{|c|}{Global Shuffle}\\
  \hline
  \hline
  Blogs & BBC & Blogs & BBC & Blogs & BBC\\
  \hline
   1.68 & 0.51 & 1.20 & 0.19 & 0.04 & 0\\
  \hline
\end{tabular}}
\caption{Mutual information for subjective probability value of subsequent comments. The value of $I(X,Y)$ was calculated with about 0.05 error due to calculation method simplifications.}\label{tab:sub}
\end{table}
\begin{figure}[!ht]
\centerline{\psfig{file=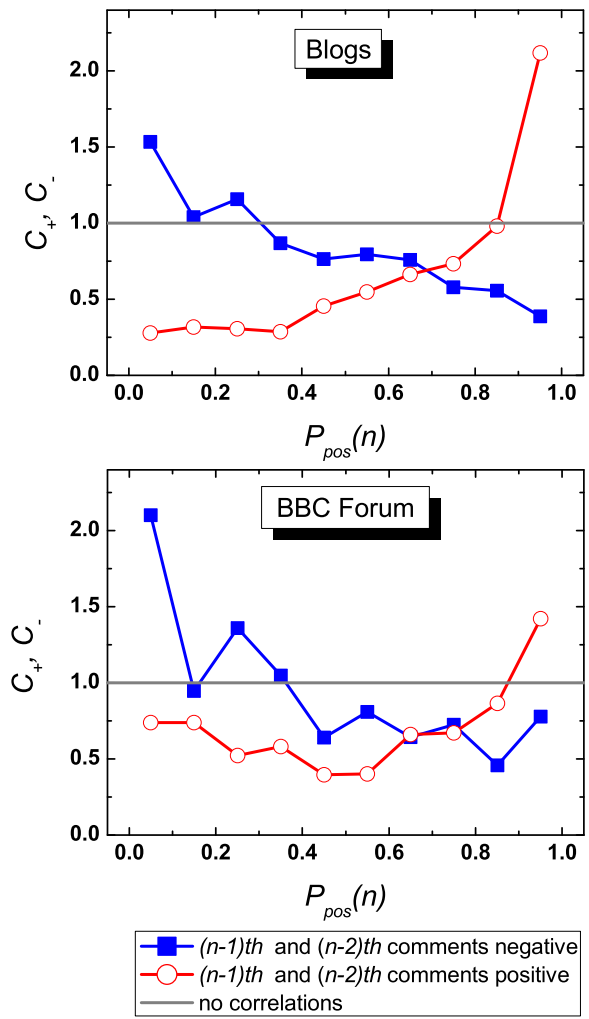,width=0.9\columnwidth}}
\caption{Three-step correlation functions $C_+$ (circles) and $C_-$ (squares) for Blogs (upper plot) and BBC Forums (bottom plot). Grey solid line indicates the level of no correlations.}
\label{fig:3step}
\end{figure}

\section{Three-step correlations}	
All subjective positive pairs and negative pairs were found in order to calculate the three-step correlation. A $P_{pos}=0.1$ binning was used so that negative comments ranged $P_{pos} \in [0,0.1]$ and positive comments ranged $P_{pos} \in [0.9,1.0]$. Owing to this it is possible to extend the previous definition of probability correlation relation (\ref{eq:c}) by 
\begin{eqnarray}
C_{+}(x_n) = \frac{p \left( x_n| x_{n-1} \ge 0.9, x_{n-2} \ge 0.9 \right)}{p(x_n)}\\
C_{-}(x_n) = \frac{p \left( x_n| x_{n-1} \le 0.1, x_{n-2} \le 0.1 \right)}{p(x_n)}
\end{eqnarray}
where $x_n$ again stands for $P_{pos}(n)$. The quantities $C_{+}(x_n)$ and $C_{-}(x_n)$ give the correlation that, after two positive (negative) posts the next one will also be positive (negative). This approach (Fig. \ref{fig:3step}) was used in order to probe the structure of subjective clusters.

Figure \ref{fig:3step} indicates that positive comments tend to group only with other positive comments. Negative comment pairs are also positively ($C_{-}(x_n) > 1$) correlated with occurrence of next negative comments, but they are also positively correlated with some unresolved comments. A common rule for both datasets could be suggested that subjective positive groups in thread tend to repel other groups.  

\section{Conclusions}	
The analysis performed on the gathered data from Internet blogs and forums shows definite signs of high correlations with the emotional content of published comments. The difference between the observed data and simulated values taken from probability distributions gives evidence of the existence of certain structures. First of all issuing a specific emotion in a comment induces with a large probability a similar emotion in the next one. This phenomenon is also seen in case of three step correlation analysis: given the fact that first two comments are highly positive/negative, the third one has a tendency to be very positive/negative as well. Such rules lead to observations of long positive, negative or objective clusters that far exceed the numbers that would have been obtained if there had been no correlations in the data. The obtained results are in agreement with our previous study showing the collective emotions and growth of emotional clusters \cite{ania_plos}.

\section{Acknowledgements}
This work was supported by a European Union grant by the 7th Framework Programme, Theme 3: Science of complex systems for socially intelligent ICT. It is part of the CyberEmotions (Collective Emotions in Cyberspace) project (contract 231323). J.S. and J.A.H. acknowledge support from Polish Ministry of Science Grant 1029/7.PR UE/2009/7 and from the European COST Action MP0801 Physics of Competition and Conflicts as well as from the Polish Ministry of Science Grant No. 578/N-COST/2009/0.

\end{document}